\documentclass[10pt,twocolumn,letterpaper]{article}

\usepackage{iccv}
\usepackage{times}

\usepackage{graphicx}
\usepackage{amsmath,amssymb}
\usepackage[bold]{hhtensor}
\usepackage{algorithm,algorithmic}
\usepackage[caption=false,font=footnotesize]{subfig}

\usepackage[pagebackref=true,breaklinks=true,colorlinks,bookmarks=false]{hyperref}

\newcommand{\squishlist}{
 \begin{list}{$\bullet$}
  { \setlength{\itemsep}{0pt}
     \setlength{\parsep}{1pt}
     \setlength{\topsep}{1pt}
     \setlength{\partopsep}{0pt}
     \setlength{\leftmargin}{1.5em}
     \setlength{\labelwidth}{1em}
     \setlength{\labelsep}{0.5em} } }

\newcommand{\squishend}{
  \end{list}  }
  
\allowdisplaybreaks

\iccvfinalcopy 

\def\iccvPaperID{478} 

\ificcvfinal\fi

\begin{document}

\title{Visual Recognition Using Directional Distribution Distance}

\author{Jianxin Wu \quad \quad \quad Bin-Bin Gao\\
National Key Laboratory for Novel Software Technology\\
Nanjing University, China\\
{\tt\small wujx2001@nju.edu.cn, gaobb@lamda.nju.edu.cn}
\and
Guoqing Liu\\
Minieye, Youjia Innovation LLC\\
{\tt\small guoqing@minieye.cc}
}

\maketitle

\begin{abstract}
In computer vision, an entity such as an image or video is often represented as a set of instance vectors, which can be SIFT, motion, or deep learning feature vectors extracted from different parts of that entity. Thus, it is essential to design efficient and effective methods to compare two sets of instance vectors. Existing methods such as FV, VLAD or Super Vectors have achieved excellent results. However, this paper shows that these methods are designed based on a generative perspective, and a discriminative method can be more effective in categorizing images or videos. The proposed D3 (discriminative distribution distance) method effectively compares two sets as two distributions, and proposes a directional total variation distance (DTVD) to measure how separated are they. Furthermore, a robust classifier-based method is proposed to estimate DTVD robustly. The D3 method is evaluated in action and image recognition tasks and has achieved excellent accuracy and speed. D3 also has a synergy with FV. The combination of D3 and FV has advantages over D3, FV, and VLAD.
\end{abstract}

\section{Introduction}

In visual recognition, an entity (object or video) is usually represented as a set of instance vectors. Each instance vector is extracted using part of the entity (\emph{e.g.}, a local window extracted from an image or a time-space subvolume extracted from a video). Various features have emerged as the state-of-the-art to extract instance vectors at different stages of recognition research, such as dense SIFT features~\cite{vi:Lowe2004}, dense CENTRIST features~\cite{me:Wu2011} or CNN features for images~\cite{vi:Jia2014}, or (improved) dense trajectory features~\cite{vi:Wang2013b} or CNN features for videos~\cite{vi:Gkioxari2015}. Although originally CNN (or other deep learning methods) integrates visual representation and classification into one system~\cite{lr:LeCun1998,vi:Krizhevsky2012}, recent works have shown that if multiple (a set of) CNN features are extracted from entities and classify images or videos based on these sets, higher accuracies can be obtained~\cite{vi:Gong2014,vi:Yoo2014,vi:Cimpoi2015,vi:Xu2015}.

Because most existing learning algorithms assume that an entity is represented as a vector instead of a set of vectors, we need to find a suitable visual representation that encodes the set of instance vectors into one single vector. It is desirable that the representation will capture useful (\emph{i.e.}, discriminative) information from the set. Thus, comparing one entity (a set of instance vectors) to another can be divided into two steps: first represent the sets as two vectors, then find a suitable distance metric to compare the vectors. One useful variant is to compare one entity to a set of entities (\emph{e.g.}, all training images, corresponding to a bigger union set by gathering the instance vectors in every image), which is often used too.

Since the $\ell_2$ distance (or correspondingly linear SVM) is very efficient and has shown great accuracy in the second step, an effective visual representation that turns a set of instance vectors into one single vector (\emph{i.e.}, the first step) has been very important in visual recognition research efforts. Many representations have been proposed, for example,
\squishlist
 \item \textbf{Fisher Vector (FV) and VLAD.} FV~\cite{vi:Sanchez2013} is based on the idea of Fisher kernel in machine learning~\cite{lr:Jaakkola1999}. It models the distribution of instance vectors in training entities using a Gaussian Mixture Model (GMM). Then, one training or testing entity is modeled generatively, by a vector which describes how the GMM can be modified to generate the instance vectors inside that entity. A GMM with $K$ components has three sets of parameters $(w_i, \vec{\mu}_i, \vec{\sigma}_i)$, $1 \le i \le K$. VLAD~\cite{vi:Jegou2012}, another popular visual representation, can be regarded as a special case of FV, by using only the $\vec{\mu}$ parameters. The classic bag-of-visual-words (BOVW)~\cite{vi:Csurka2004} representation is also a special case of FV, using the $w$ parameters.
 \item \textbf{Super-Vector} Instead of modeling the instance vectors as distributions, the Super-Vector~\cite{vi:Zhou2010} represents a set of instance vectors based on how they can be reconstructed from dictionary items. Super-Vector aims at reducing the reconstruction error, which is also from a generative perspective. The output of Super-Vector has two parts, which are conceptually related to the $w$ and $\vec{\mu}$ parameters in Fisher Vectors.
\squishend

Both threads of methods have shown excellent accuracy in the literature. However, they both focus on modeling how one entity or one distribution is \emph{generated}. Given the fact that the task in hand is recognition, we argue that \emph{we need to pay more attention to how two entities or two distributions are separated.} In other words, we need a visual representation that pays more attention to the discriminative side. We naturally expect that such a representation would be suitable for visual recognition tasks, whose objective is to properly separate entities belonging to different categories.

In this paper, we propose a discriminative distribution distance (D3) representation that converts a set of instance vectors into a vector representation. D3 explicitly considers two distributions: a density $X$ which is estimated from the training set as a reference model, and one entity forms another distribution $Y$. D3 then uses the distribution distance between $X$ and $Y$ as a discriminative representation for the entity $Y$. Technically, D3 has the following contributions.
\squishlist
 \item We propose a direction total variation distance (DTVD) to measure the distance between $X$ and $Y$, which contains more discriminative information than classic distribution distances by considering \emph{directions};
 \item Directly calculating DTVD is unstable and problematic because $Y$ may be non-Gaussian and only contains few items. We propose to estimate DTVD in a discriminative manner, by \emph{calculating robust classification errors} when we try to classify every dimension of $X$ from $Y$;
 \item We also show that D3 and FV are complementary to each other. By combining D3 and FV, we can achieve an accuracy higher than D3, FV, and VLAD.
\squishend

We will start by explaining closely related methods, then proposing the directional distribution distance, its robust estimation, and the entire D3 pipeline in Sec.~\ref{sec:D3}. Sec.~\ref{sec:results} presents empirical results, and Sec.~\ref{sec:conclusions} concludes this paper.

\section{Discriminative Distribution Distance} \label{sec:D3}

In this section, we propose a discriminative distribution distance (abbreviated as D3) to compare two sets of observations, which leads to an efficient and effective visual representation.

\subsection{Distribution distance: generative vs. discriminative}

Given two objects $X$ and $Y$, each of which is represented as an unordered set of instance vectors, \emph{i.e.}, $X=\{\vec{x}_1, \ldots, \vec{x}_{n_X}\}$, $Y = \{\vec{y}_1, \ldots, \vec{y}_{n_Y}\}$, we are interested in finding $d(X,Y)$, the distance (or dissimilarity) between them. This task is frequently encountered in compute vision. For example, an image or a video is usually represented as a set of feature vectors extracted from various image patches or supervoxels.

In the Fisher Vector (FV) representation, a large set of instance vectors are extracted from training images or videos. We treat this set as $X$ and a Gaussian Mixture Model (GMM) $p_X$ with parameters $\lambda = \bigl\{(w_k,\vec{\mu}_k,\vec{\Sigma}_k)\bigr\}_{k=1}^K$ is estimated from $X$. When a test image or video $\mathcal{Y}$ is presented, we extract its instance vectors and treat it as $Y$. The FV representation considers $X$ and $Y$ as generated from two underlying distributions $p_X$ and $p_Y$, and encodes $\mathcal{Y}$ as a vector $\vec{f}$. This is a generative model and each component in $\vec{f}$ describes how each parameter in $\lambda$ should be modified such that $p_X$ can be modified to fit the data $Y$ properly.

Specifically, the probability that $\vec{y}_i$ is generated by the $k$-th Gaussian is
\begin{equation}
 \gamma_i(k) = p(k|\vec{y}_i,\lambda) = \frac{1}{Z} w_k p_k(\vec{y}_i|\lambda)\,, 
\end{equation}
where $Z$ is a normalization constant, $p_k$ is the $k$-th Gaussian component with weight $w_k$, and parameters $(\vec{\mu}_k,\Sigma_k)$. In FV, the GMM covariances $\Sigma_k$ are assumed to be diagonal, whose diagonal entries form a vector $\vec{\sigma}_k$. The trends of parameter changing (gradients) that modify $p_X$ to fit $\mathcal{Y}$ is then (for all $1 \le k \le K$)~\cite{vi:Sanchez2013}
\begin{align}
 \vec{f}_{w_k}         &= \frac{1}{\sqrt{w_k}} \sum_{i=1}^{n_Y} \left( \gamma_i(k) - w_k \right) \,, \\
 \vec{f}_{\vec{\mu}_k} &= \frac{1}{\sqrt{w_k}} \sum_{i=1}^{n_Y} \gamma_i(k) \left( \frac{\vec{y}_i - \vec{\mu}_k}{\vec{\sigma}_k} \right) \,, \\
 \vec{f}_{\vec{\sigma}_k} &= \frac{1}{\sqrt{2w_k}} \sum_{i=1}^{n_Y} \gamma_i(k) \left( \frac{(\vec{y}_i - \vec{\mu}_k)^2}{(\vec{\sigma}_k)^2} - 1 \right) \,,
\end{align}
which correspond to the $w, \vec{\mu}, \vec{\sigma}$ parameters for $p_X$, respectively. The image or video $\mathcal{Y}$ is then represented by a feature vector $\vec{f}$, which concatenates $\vec{f}_{w_k}$, $\vec{f}_{\vec{\mu}_k}$, and $\vec{f}_{\vec{\sigma}_k}$ for all $1 \le k \le K$.

We want to emphasize two observations based on the FV representation.
\squishlist
 \item The gradient vector $\vec{f}$ is formed under the \emph{generative} assumption that $Y$ can be modeled by $p_X$ if we are allowed to modify the parameter set $\lambda$. Since what we are really interested in is how far is $X$ from $Y$, we believe that a \emph{discriminative} distance between $X$ and $Y$ is a better option. That is, in this paper we will treat $X$ and $Y$ as sampled from two different distributions $p_X$ and $p_Y$, and find their distribution distance to encode the image or video $\mathcal{Y}$;
 \item Since diagonal $\Sigma_k$ are used, after the soft assignment probabilities $\gamma_i(k)$ are calculated, each dimension of $\vec{f}$ is generated independent of any other dimension. Thus, in finding a suitable representation for $Y$, we only need to consider each dimension individually. The problem is then: given two sets of \emph{scalar} values $X=\{x_1,x_2,\ldots,\}$ and $Y=\{y_1,y_2,\ldots\}$ (sampled from 1-d distribution $p_X$ and $p_Y$, respectively), how do we properly compute $d(p_X,p_Y)$?
\squishend

As a final note in this section, the VLAD and super vector representation can be interpreted as special cases of FV, while VLAD uses the $\vec{f}_{\vec{\mu}_k}$ components of $\vec{f}$, and super vectors use both $\vec{f}_{w_k}$ and $\vec{f}_{\vec{\mu}_k}$. It is also a common practice to use only $\vec{f}_{\vec{\mu}_k}$ and $\vec{f}_{\vec{\sigma}_k}$ in FV implementations.

\subsection{Directional Total Variation Distance}

We need to be more discriminative. Thus, we propose to explicitly consider two distributions $X$ and $Y$, where $X$ is a density of instance vectors estimated from the training set, and $Y$ is from one (training or testing) entity. A representation of $Y$ that encodes the \emph{distance} between $X$ and $Y$ will contain useful discriminative information about $Y$.

\begin{figure}[t]
 \centering
 \subfloat[]
 {
   \includegraphics[width=0.375\textwidth]{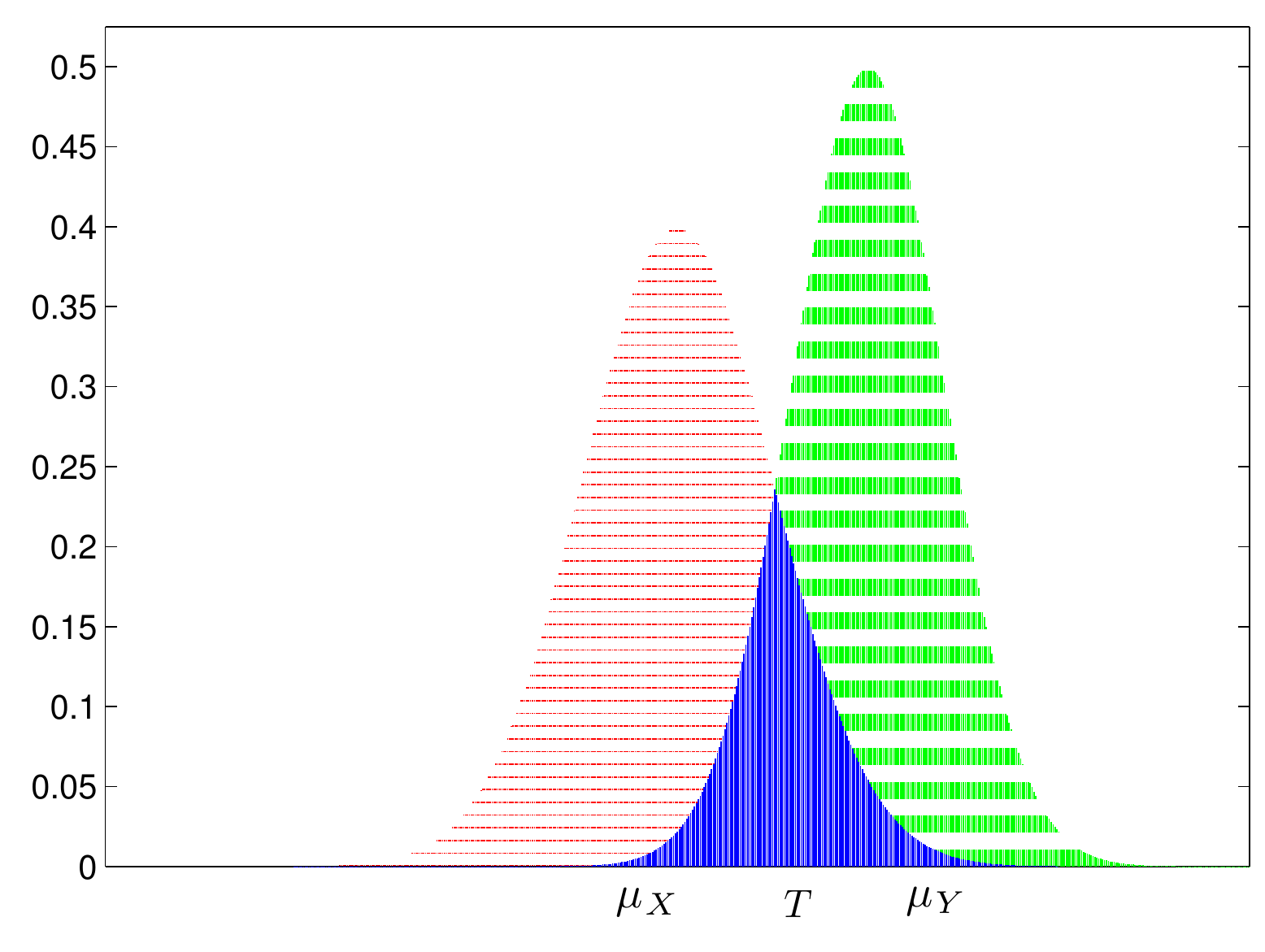} \label{fig:tvd_area}
 }
 \hspace{8pt}
 \subfloat[]
 {
   \includegraphics[width=0.375\textwidth]{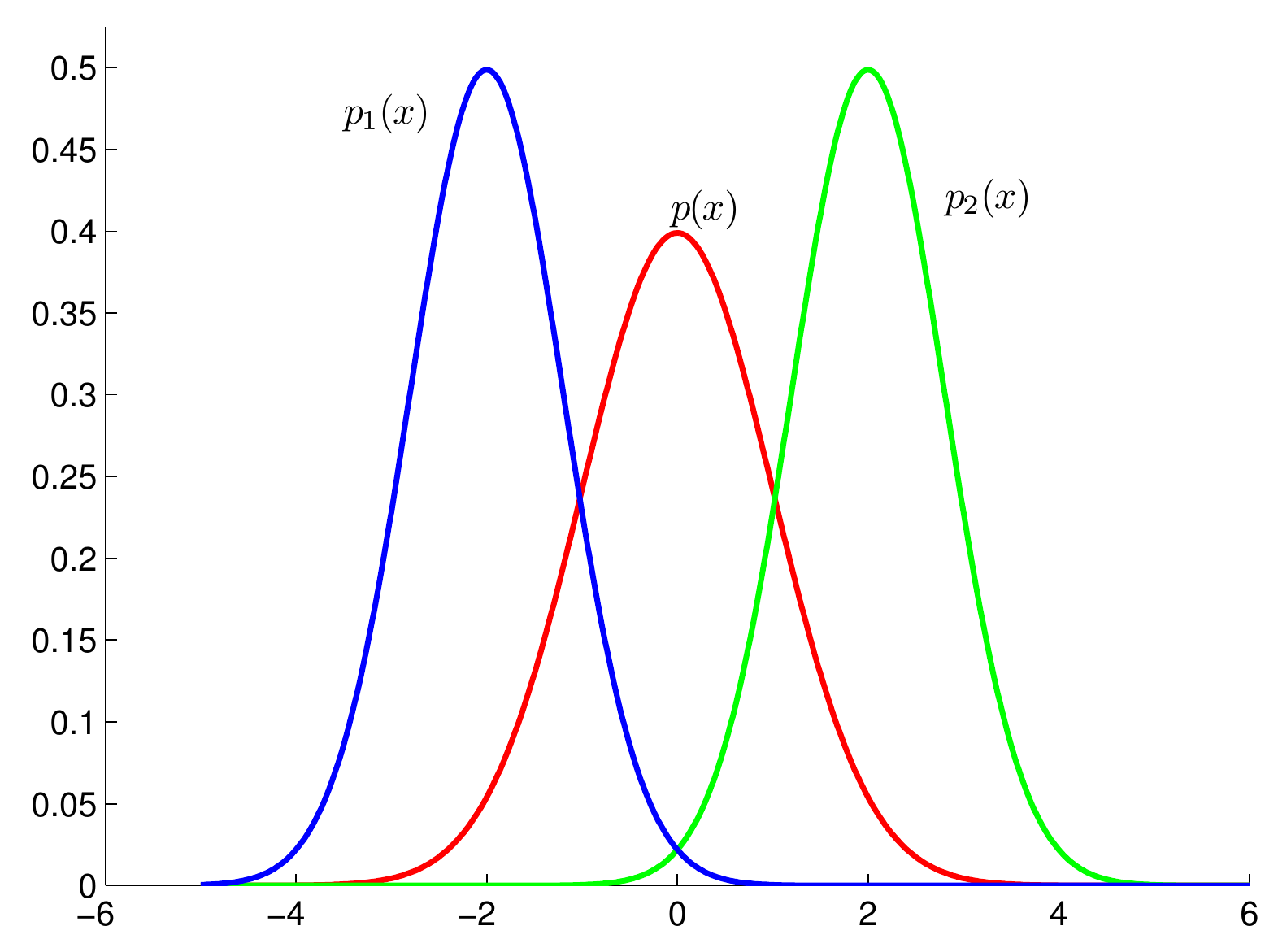} \label{fig:tvd_directional}
 }
 \caption{Illustration of the total variation distance. \ref{fig:tvd_area} illustrates $d_{TV}$ for two Gaussians, and~\ref{fig:tvd_directional} reveals that \emph{direction} is essential.} \label{fig:tvd}
\end{figure}

A widely used distance that compares two distributions is the \emph{total variation} distance, which is independent of the distributions' parameterizations. Let $\nu_1$ and $\nu_2$ be two probability measures on a measurable space $(\mathcal{O},\mathcal{B})$, the total variation distance is defined as
\begin{equation}
 d_{TV}(\nu_X,\nu_Y) \triangleq \sup_{A \in \mathcal{B}} |\nu_X(A)-\nu_Y(A)| \,. \label{eqn:tvd}
\end{equation}
While this definition is rather formal, $d_{TV}$ has a more intuitive equation for commonly used continuous distributions by the Scheffe's Lemma~\cite{lr:Dasgupta2011}. For example, for two normal distributions with p.d.f. $X \sim N(\mu_X,\sigma_X^2)$ and $Y \sim N(\mu_Y,\sigma_Y^2)$,
\begin{equation}
 d_{TV}(p_X,p_Y) = \frac{1}{2} \int_u |p_X(u)-p_Y(u)| \,\mathrm{d}u \,. \label{eqn:tvd_normal}
\end{equation}
As illustrated in Fig.~\ref{fig:tvd_area}, it is half the summed area of the red and green regions, which clearly indicates how two distributions are separated from each other.

The classic total variation distance (Eq.~\ref{eqn:tvd_normal}), however, is missing one most important information that captures the key difference between $p_X$ and $p_Y$, as shown in Fig.~\ref{fig:tvd_directional}. In Fig.~\ref{fig:tvd_directional}, $p_1$ and $p_2$ are symmetric with respect to the mean of $p$, thus we have $d_{TV}(p,p_1)=d_{TV}(p,p_2)$, in spite of the fact that $p_1$ and $p_2$ are far apart. The missing of directional information is responsible for this drawback. Thus, we propose a directional total variation distance (DTVD) as
\begin{equation}
 d_{DTV}(p_X,p_Y) = \mathrm{sign}(\mu_Y-\mu_X) \times d_{TV}(p_X,p_Y) \,. \label{eqn:dtv}
\end{equation}
DTVD is a signed distance. In Fig.~\ref{fig:tvd_directional}, we will (correctly) have $d_{DTV}(p,p_1) = - d_{DTV}(p,p_2)$, which clearly signifies the difference between $p_1$ and $p_2$. Obviously the distance function $d_{DTV}$ is not a metric, because it is neither non-negative, nor symmetric.

\subsection{Robust estimation of the DTVD}

\begin{figure}[t]
 \centering
 \subfloat[]
 {
   \includegraphics[width=0.225\textwidth]{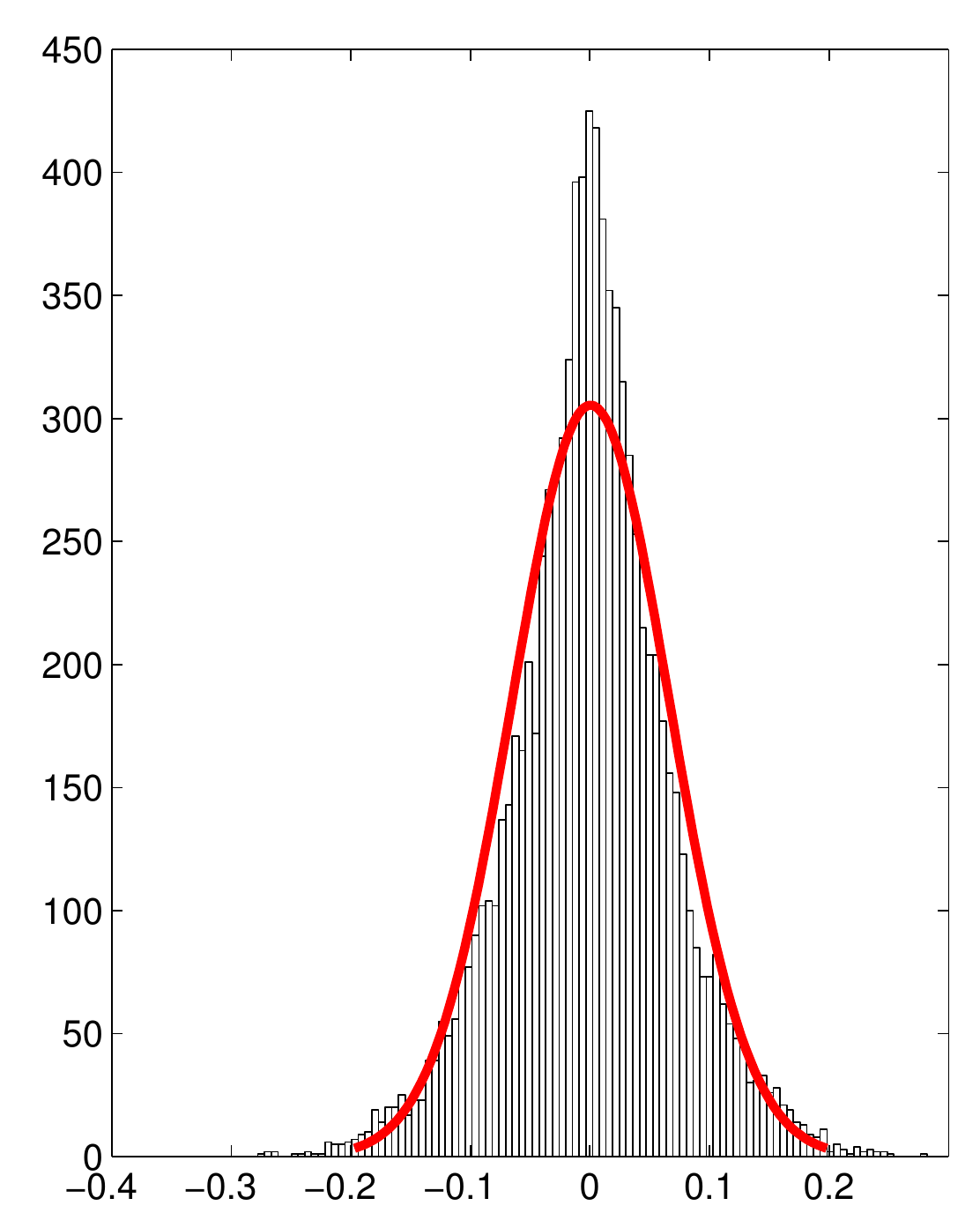} \label{fig:37th_codebook}
 }
 \subfloat[]
 {
   \includegraphics[width=0.225\textwidth]{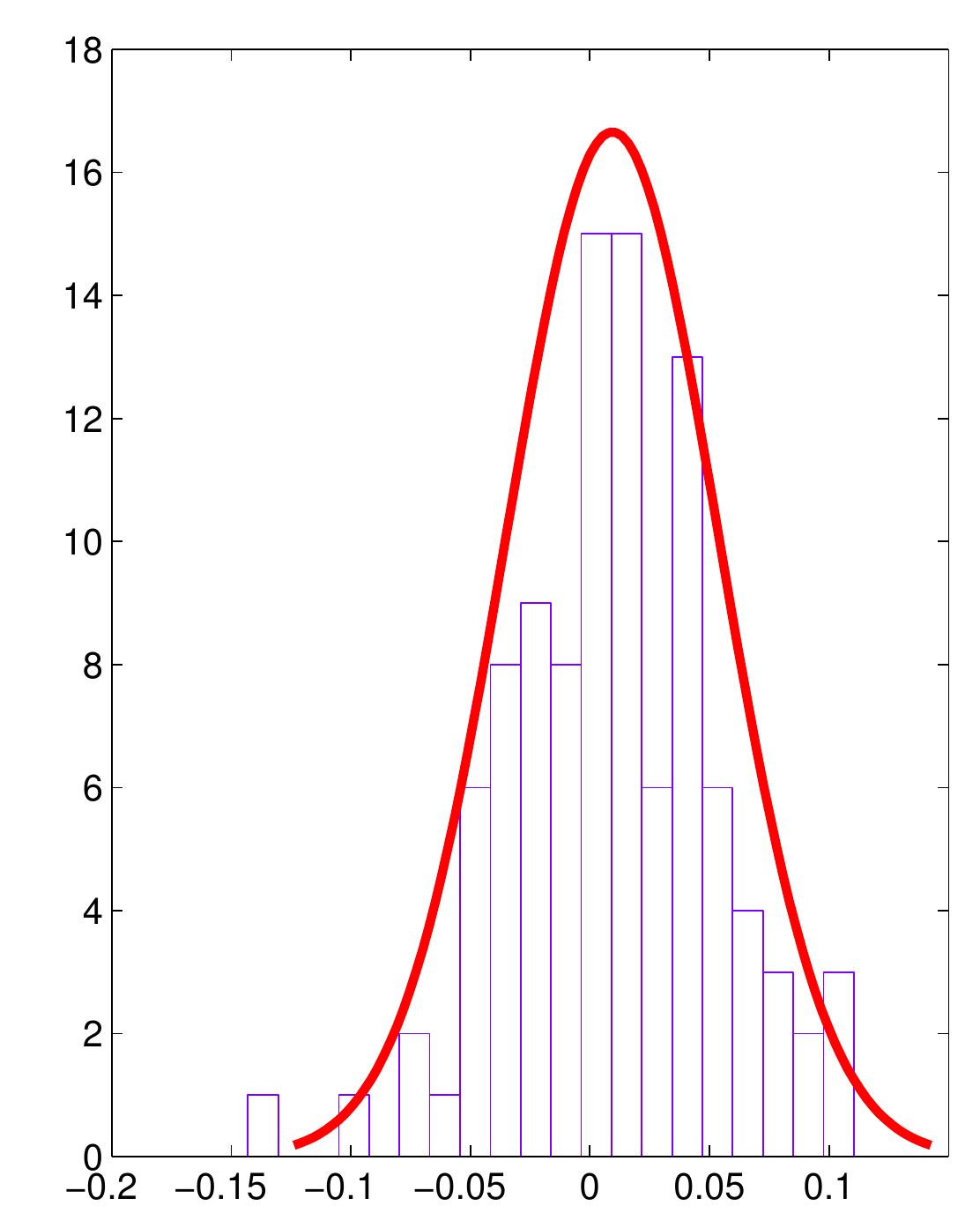} \label{fig:37th_image}
 }
 \caption{Typical distribution of feature values. \ref{fig:37th_codebook} is calculated based on features used to generate the codebook, and~\ref{fig:37th_image} is from a single image. The red curve is a normal distribution estimated from the same data. This figure is generated using bag of dense SIFT on the Scene 15 dataset with $K=64$ VLAD encoding. The dimension shown is the 37-th dimension in the 37-th cluster of the codebook.} \label{fig:37th}
\end{figure}

For two Gaussians $p_X$ and $p_Y$, their p.d.f. will have two intersections if $\sigma_X \neq \sigma_Y$. For example, in Fig.~\ref{fig:tvd_area} the second intersection is in the far right end of the $x$-axis. A closed-form solution to calculate $d_{TV}$ based on both intersections is available~\cite{lr:Dasgupta2011}. However, this closed-form solution leads to serious performance drop when used in visual recognition in our experiments. We conjecture that two reasons have caused this issue:
\squishlist
 \item The distributions are not necessarily normal. As shown in Fig.~\ref{fig:37th}, the typical example of $p_X$ (in Fig.~\ref{fig:37th_codebook}) is generated from many training instances, its shape resembles that of a Gaussian, but has a shaper peak. $p_Y$, which is generated from a single image, deviates from a normal distribution;
 \item Since the set $Y$ (which is extracted from a single image or video, cf. Fig.~\ref{fig:37th_image}) usually contains small number of instance vectors, this fact leads to unstable estimation of its distribution parameters, and hence unstable $d_{DTV}(p_X,p_Y)$. Thus, we need a more robust way to estimate the distribution distance.
\squishend

Our key insight again arises from the discriminative perspective. It is obvious that the total variation distance $d_{TV}$ is equivalent to one minus the \emph{Bayes error} of a binary classification problem, where the two classes have equal prior and follow $p_X$ and $p_Y$, respectively. Thus, we can estimate $d_{TV}$ (hence $d_{DTV}$) by \emph{robustly estimating the classification error between the two sets of examples $X$ and $Y$}. Note that this task is easy since $X$ and $Y$ only contain scalar examples.

We adopt the minimax probability machine (MPM)~\cite{lr:Lanckriet2002} to estimate the classification error. MPM is robust because it minimizes the maximum probability of misclassification, hence the name minimax. Given examples $X$ with mean $\vec{\mu}_X$ and covariance $\Sigma_X$ and examples $Y$ with $\vec{\mu}_Y$ and $\Sigma_Y$, the classifier boundary $\vec{a}^T\vec{x}-b=0$ is determined by the MPM problem
\begin{align}
 \kappa_\star^{-1} & \triangleq \min_{\vec{a}} \sqrt{\vec{a}^T\Sigma_X \vec{a}} + \sqrt{\vec{a}^T\Sigma_Y \vec{a}} \\
 \text{s.t. } & \;\;\; \vec{a}^T(\vec{\mu}_X-\vec{\mu}_Y) = 1 \;, \label{eqn:mpm}
\end{align}
and
\begin{equation}
 b_\star = \vec{a}_\star^T \vec{\mu}_X - \kappa_\star\sqrt{\vec{a}_\star^T \Sigma_X \vec{a}_\star} \,.
\end{equation}

Eq.~\ref{eqn:mpm} is a second order cone problem (SOCP) that can be solved by an iterative algorithm. However, since we are dealing with scalar examples that (assumed to) follow normal distributions, it has a closed form solution. Note that $X \sim N(\mu_X,\sigma_X^2)$ and $Y \sim N(\mu_Y,\sigma_Y^2)$, we can immediately get the following boundary $a_\star x-b_\star=0$, where
\begin{align}
 a_\star &= \frac{1}{\mu_X-\mu_Y}  \,, \;\; b_\star = a_\star \times \frac{\mu_X\sigma_Y+\mu_Y\sigma_X}{\sigma_X+\sigma_Y} \,, \\
 \kappa_\star &= \frac{|\mu_X-\mu_Y|}{\sigma_X+\sigma_Y}  \,.
\end{align}
That is, the two 1-d distributions $p_X$ and $p_Y$ are classified at the threshold value
\begin{equation}
  T = \frac{\mu_X\sigma_Y+\mu_Y\sigma_X}{\sigma_X+\sigma_Y} \,. \label{eqn:T}
\end{equation}

If we re-use Fig.~\ref{fig:tvd_area} and (approximately) assume the red, blue, and green areas intersect at $T=\frac{\mu_X\sigma_Y+\mu_Y\sigma_X}{\sigma_X+\sigma_Y}$, which is guaranteed to reside in between $\mu_X$ and $\mu_Y$. Then, the area of the blue region is:
\begin{equation}
 Area = 1 - \Phi\left(\frac{T-\mu_X}{\sigma_X}\right) + \Phi\left(\frac{T-\mu_Y}{\sigma_Y}\right) \,.
\end{equation}
where 
\begin{equation}
 \Phi(x)=\frac{1}{\sqrt{2\pi}}\int_{-\infty}^x e^{-t^2/2} \,\mathrm{d}t 
\end{equation}
is the cumulative distribution function (c.d.f.) of a standard normal distribution $N(0,1)$. And, we have
\begin{equation}
 d_{DTV}(p_X,p_Y) = 2 - 2Area = 4\Phi \left( \frac{\mu_Y-\mu_X}{\sigma_X+\sigma_Y} \right) - 2 \,, \label{eqn:dtv2}
\end{equation}
making use of the fact that \[ \frac{T-\mu_X}{\sigma_X}=\frac{\mu_Y-\mu_X}{\sigma_X+\sigma_Y}=-\frac{T-\mu_Y}{\sigma_Y}, \] and the property of $\Phi$ that $\Phi(-x)=1-\Phi(x)$.

Two points are worth mentioning about Eq.~\ref{eqn:dtv2}.
\squishlist
 \item Although our derivation and Fig.~\ref{fig:tvd_area} is assuming $\mu_X<\mu_Y$, it is easy to derive that when $\mu_X \ge \mu_Y$, Eq.~\ref{eqn:dtv2} still holds. And, it always have the same sign as $\mu_Y-\mu_X$. Hence, Eq.~\ref{eqn:dtv2} computes $d_{DTV}$ instead of $d_{TV}$, and its range is $[-2 \;\; 2]$.
 \item In practice we use the error function. The error function is defined as 
 \begin{equation}
  \mathrm{erf}(x)=\frac{1}{\sqrt{\pi}}\int_{-x}^x e^{-t^2} \,\mathrm{d}t \,,
 \end{equation}
and it satisfies that
\begin{equation}
 \Phi(x)=\frac{1}{2}\left(1+\mathop{\mathrm{erf}}\left(\frac{x}{\sqrt{2}}\right)\right) \,.
\end{equation}
Thus, we have
 \begin{equation}
  d_{DTV}(p_X,p_Y) = 2\mathop{\mathrm{erf}}\left( \frac{\mu_Y-\mu_X}{\sqrt{2}(\sigma_X+\sigma_Y)} \right) \,. \label{eqn:dtv3}
 \end{equation}
   The error function $\mathrm{erf}$ is built-in and efficient in most major programming languages, which facilitates the calculation of $d_{DTV}$ using Eq.~\ref{eqn:dtv3}.
\squishend

We also want to note there has been research to model the discriminative distance between two sets of instance vectors. In~\cite{vi:Poczos2012}, non-parametric kernels are estimated from two sets, and use the Hellinger's distance or the R\'{e}nyi-$\alpha$ divergence to measure the distance between two distributions. This method, however, suffers from one major limitation. Non-parametric kernel estimation is very time consuming, which took 3.3 days in a subset of the Scene 15 dataset, a fact that renders it impractical for large problems. As a direct comparison, D3 only requires less than 2 minutes.

\subsection{The pipeline using $d_{DTV}$ for visual recognition}

We assume that an image or video $\mathcal{Y}$ is represented as a bag of instance feature vectors $Y=\{\vec{y}_1,\vec{y}_2,\ldots\}$, where each $\vec{y}_i \in \mathbb{R}^d$. The instance vectors are usually extracted as dense SIFT vectors or deep learning (CNN) features for images, or dense trajectory features or deep learning features for videos, or other representations that use a set of vectors to represent an entity.

The pipeline to use $d_{DTV}$ to generate image or video representation follows two steps.
\squishlist
 \item \textbf{Dictionary generation.} For simplicity and computational efficiency, we collect a large set of instance vectors from the training set, and then use the $k$-means algorithm to generate a dictionary that partitions the space of instance vectors into $K$ regions. We compute the mean and standard deviation of the instance vectors inside cluster $k$ as $\vec{\mu}_k$ and $\vec{\sigma}_k$ for all $1 \le k \le K$. Values in the standard deviation vector $\vec{\sigma}_k$ is computed for every dimension independently;
 \item \textbf{Visual representation.} Given an image or video $\mathcal{Y}$, we use Algorithm~\ref{alg:encoding} to convert it to a vector representation. Note that since we normalize every $\vec{f}_i$ in Algorithm~\ref{alg:encoding}, the constant factor (`2') in Eq.~\ref{eqn:dtv3} is not necessary and is thus omitted.
\squishend

\begin{algorithm}[t]
 \caption{Visual representation using D3} \label{alg:encoding}
 \begin{algorithmic}[1]
  \STATE \textbf{Input}: An image or video $Y=\{\vec{y}_1,\vec{y}_2,\ldots\}$; and, \\ a dictionary (visual code book) with size $K$ and cluster mean $\vec{\mu}_k$ and standard deviations $\vec{\sigma}_k$ ($1 \le k \le K$)
  \FOR{$i=1,2,\ldots,K$}
    \STATE $Y' = \{ \vec{y}_j | \vec{y}_j \in Y, \mathop{\arg\min}_{1 \le k \le K}\|\vec{y}_j - \vec{\mu}_k \|=i \}$
    \STATE Compute the mean and standard deviation vectors of the set $Y'$, denote as $\vec{\mu}'$ and $\vec{\sigma}'$, respectively
    \STATE $\vec{f}_i = \mathop{\mathrm{{erf}}}\left( \frac{\vec{\mu}' - \vec{\mu}_i}{\sqrt{2}(\vec{\sigma}' + \vec{\sigma}_i)} \right)$. \\Note that the $\mathrm{erf}$ function is applied to every component of the $\frac{\vec{\mu}' - \vec{\mu}_i}{\sqrt{2}(\vec{\sigma}' + \vec{\sigma}_i)}$ vector individually
    \STATE $\vec{f}_i = \frac{\vec{f}_i}{\|\vec{f}_i\|}$
  \ENDFOR
  \STATE $\vec{f} = [\vec{f}_1^T \; \vec{f}_2^T \; \ldots \; \vec{f}_K^T]$
  \STATE $\vec{f} = \frac{\vec{f}}{\|\vec{f}\|}$
  \STATE \textbf{Output}: The new representation $\vec{f} \in \mathbb{R}^{d \times K}$ 
 \end{algorithmic}
\end{algorithm}

In Algorithm~\ref{alg:encoding}, we use the $k$-means algorithm to generate a visual codebook, and an instance vector is hard-assigned to one visual code word. A GMM model can also be used as a soft codebook, similar to what is performed in FV. However, a GMM has higher costs in generating both the dictionary and visual representation. Thus, we use $k$-means to generate a codebook in D3. Then, D3 and VLAD have very similar frameworks, and it is interesting to compare D3 with both VLAD and FV.

\subsection{Efficiency and hybrid representation}

Since the error function implementation is efficient, the computational cost of D3 is roughly the same as that of VLAD, which is much more efficient than the FV method. The evaluation in~\cite{vi:Peng2014} showed that the time for VLAD is only less than 5\% of that of FV. Thus, a visual representation using D3 is efficient to compute.

It is also worth noting that although D3 and FV both used first- and second-order statistics of an image or video $Y$ and compare these statistics with those computed from the training set, they use these statistics in very different ways. Thus, different information (discriminative vs. generative) are extracted by D3 and FV. By computing the D3 and FV representation separately and then concatenate them together to form a hybrid one, we can get higher recognition accuracy than both D3 and FV, as will be shown in Sec.~\ref{sec:results}. Suppose we form a $dK$ dimensional D3 vector and a $dK$ dimensional FV vector, the hybrid representation will be $2dK$ dimensional. However, its computational time will be only roughly half of that of forming a $2dK$ dimensional FV representation.

A final note is about higher order VLAD. VLAD only uses first-order statistics (mean) of the set of instance vectors. In~\cite{vi:Peng2014}, higher-order statistics (variance and skewness) are added to effectively improve VLAD. Since this method will triple the number of dimensions of VLAD (with the same $K$) and its accuracy is not as high as FV, we will not empirically compare D3 with this method in this paper. However, because D3 does not specify how a codebook is generated, the supervised codebook generation method of~\cite{vi:Peng2014} can be adopted to further improve D3 in the future.

\section{Experimental Results} \label{sec:results}

To compare the representations fairly, we compare them using the same number of dimensions. For example, the following setups will be compared to each other.
\squishlist
 \item D3 (or VLAD) with $K_1=256$ visual words; the representation has $dK_1$ dimensions;
 \item FV with $K_2=128$ components ($2dK_2=dK_1$);
 \item A mixture of D3 and FV with $K_3=128$ in D3 and $K_4=64$ in FV ($dK_3 + 2dK_4=2dK_2=dK_1$).
\squishend
We will use D3's $K$ size to indicate the size of all the above setups (\emph{i.e.}, $K=256$ in this example).

Three types of experiments are performed. First, a small image dataset is used to study the property of D3 (Sec.~\ref{sec:exp_property}). Then, D3 is evaluated in action recognition (with the ITF features, in Sec.~\ref{sec:exp_action}) and in image recognition (with CNN features, in Sec.~\ref{sec:exp_image_deep}). Discussions are in Sec.~\ref{sec:exp_discussions}.

\subsection{Why use D3?} \label{sec:exp_property}

We first study the properties of the proposed D3 representation, and shed some lights on why it is an effective way to encode the distance between two sets of dense SIFT instance vectors. 

Using the training images of the Scene 15 dataset~\cite{pl:Lazebnik2006} and dense SIFT features (with step size 4), we compare the per-dimensional discriminative power of these two representations (D3 and VLAD).

\begin{figure}[t]
 \centering
 \subfloat[]
 {
   \includegraphics[width=0.225\textwidth]{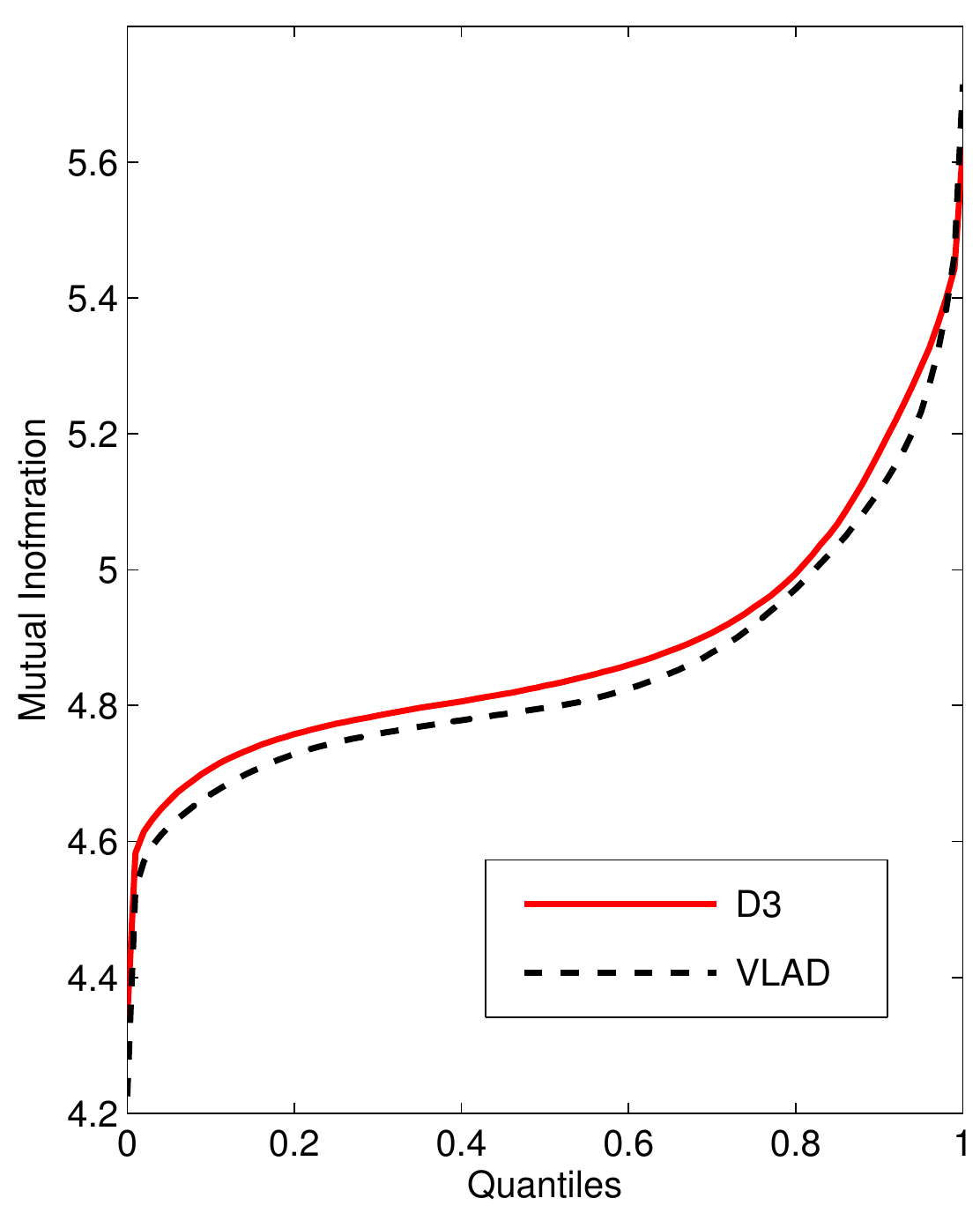} \label{fig:MI_quantile}
 }
 \subfloat[]
 {
   \includegraphics[width=0.225\textwidth]{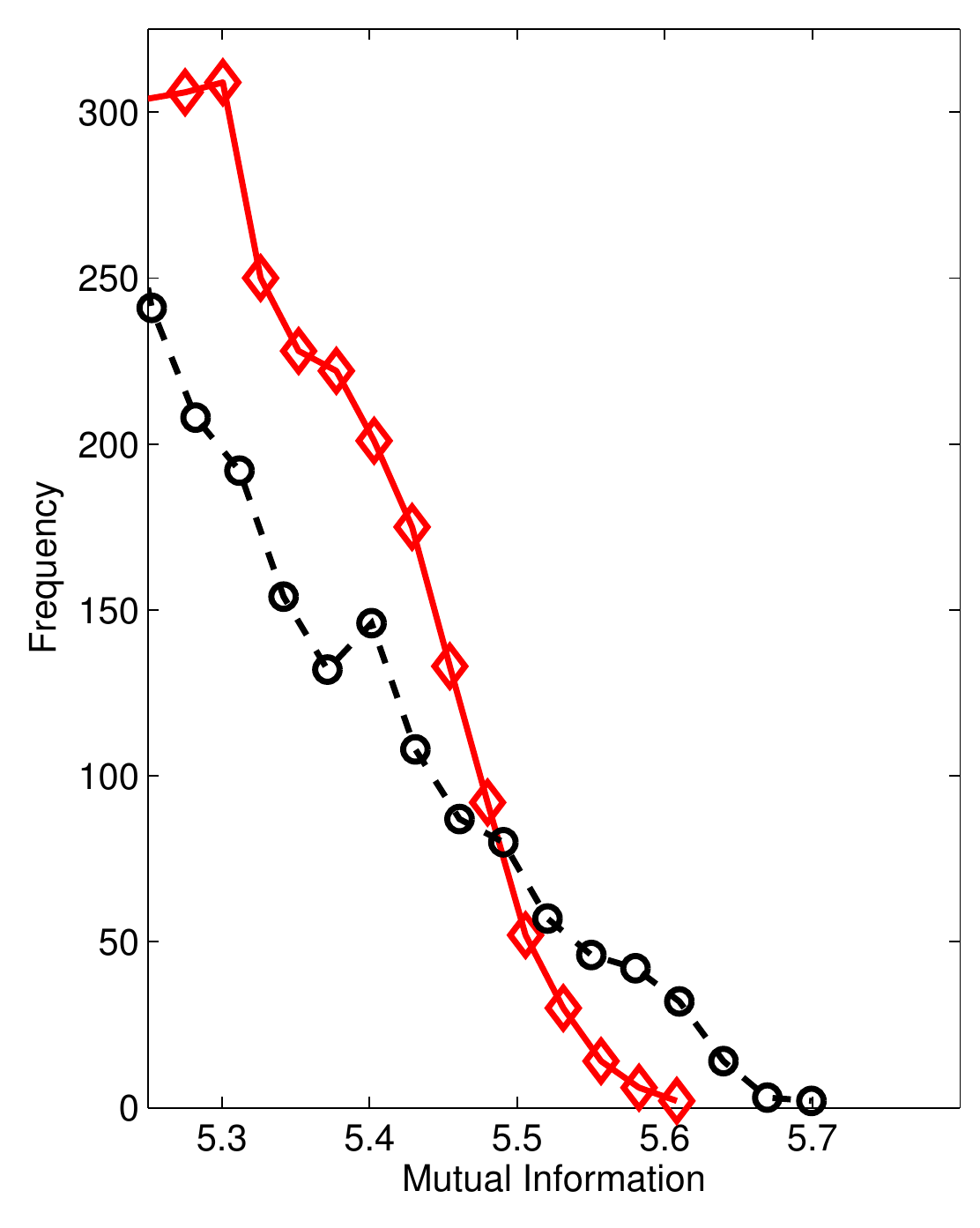} \label{fig:MI_top}
 }
 \caption{Distribution of per-dimensional mutual information. \ref{fig:MI_quantile} shows the quantile values in the full range, and~\ref{fig:MI_top} is the frequency of high (most discriminative) MI values.} \label{fig:MI}
\end{figure}

Suppose $X$ is the D3 or VLAD representation of a set of images with corresponding image labels $\vec{l}$, whose $i$-th dimension form a vector $\vec{x}_{:i}$. It is natural to measure the discriminative power of the $i$-th dimension by computing the mutual information between $\vec{x}_{:i}$ and $\vec{l}$, \emph{i.e.}, $\mathrm{MI}(\vec{x}_{:i},\vec{l})$~\cite{me:Wu2014_choose}. We use the 2-bit method in~\cite{me:Wu2014_choose} to quantize $\vec{x}_{:i}$ and compute the mutual information. The distribution of all dimension's MI values are shown in Fig.~\ref{fig:MI}.

Fig.~\ref{fig:MI_quantile} shows the MI's quantile values. For example, when the $x$-axis is 0.5, the D3 and VLAD curve has value 4.8292 and 4.7966, meaning that the median of D3's MI value is above the of VLAD's by 0.0326. Similarly, when the $y$-axis is 4.8282, the D3 and VLAD curves has corresponding quantile ($x$-axis) values 0.5 and 0.3888, meaning that 50\% of D3's MI value is above 0.4282, but only 38.88\% of VLAD's dimensions reaches this discriminative power. Since the D3 (red solid) curve is almost consistently above the VLAD (dashed black) curve, D3's dimensions have higher discriminative power than VLAD's.

Fig.~\ref{fig:MI_top} shows the frequencies of dimensions that have the highest MI values. Although VLAD has a few dimensions that have higher MI values than D3, D3 obviously have many more discriminative dimensions. The total number of dimensions with MI values $>5.25$ are 2020 and 1544 for D3 and VLAD, respectively, a 30.8\% advantage for D3. Since every single dimension is too weak to classify the image, it is more important to have many dimensions with good discriminative powers than having few only slightly more discriminative ones.

\subsection{Action recognition results} \label{sec:exp_action}

We first show experimental results for action recognition. A set of improved trajectory features (ITF)~\cite{vi:Wang2013b} are extracted and then converted to D3, VLAD, FV, and two hybrid representations (D3+FV and VLAD+FV). The default parameters are used to extract ITF features.

We experimented on three datasets: UCF 101~\cite{vi:UCF101}, HMDB 51~\cite{vi:Kuehne2011} and Youtube~\cite{vi:Liu2009}. For UCF 101, the three splits of train and test videos in~\cite{vi:THUMOS2013} are used and we report the average accuracy. This dataset has 13320 videos and 101 action categories. The HMDB 51 dataset has 51 actions in 6766 clips. We use the original (not stabilized) videos and follow~\cite{vi:Kuehne2011} to report average accuracy of its 3 predefined splits of training / testing videos. Youtube is a small scale dataset with 11 action types. There are 25 groups in each action category and 4 videos are used in each group. Following the original protocol, we report the average of the 25-fold leave one group cross validation accuracy rates. Results on these datasets are reported in Table~\ref{tbl:action}. We summarize the experimental results into the following observations.

\begin{table*}
 \caption{Action recognition accuracy (\%) comparisons. Note that the results in one column are compared with the same number of dimensions in the representations. For example, the column $K$=256 means that $K=256$ for D3 and VLAD, $K=128$ for FV, and in the hybrid representation, $K=128$ for D3 or VLAD combined with $K=64$ for FV. Note that $K=64$ results for the hybrid representation is not presented. The best results are shown in bold face.} \label{tbl:action}
 \centering
 \small
 \begin{tabular}{| l | *{4}{r} || *{4}{r} || *{4}{r} |}
  \hline
            & \multicolumn{4}{c||}{UCF 101} & \multicolumn{4}{c||}{HMDB 51} & \multicolumn{4}{c|}{Youtube} \\ 
\multicolumn{1}{|r|}{$K$} & 512   & 256   & 128   & 64    & 512   & 256   & 128   & 64    & 512   & 256   & 128   & 64 \\ \hline
      D3                  & 84.35 & 84.32 & 83.03 & 81.34 & 56.14 & 55.29 & 54.71 & 51.70 & 89.91 & 91.55 & 91.09 & 90.36  \\ \hline
      VLAD                & 82.81 & 82.54 & 81.59 & 79.78 & 55.45 & 55.14 & 53.92 & 50.22 & 90.00 & 89.73 & 89.18 & 89.09  \\ \hline 
      FV                  & 85.23 & 84.82 & 83.80 & 82.48 & 58.13 & 57.34 & 55.88 & 53.20 & 91.00 & 91.00 & 90.73 & 90.45  \\ \hline \hline
      D3+FV               & {\bf 85.92} & {\bf 85.44} & {\bf 84.20} &       & {\bf 58.34} & {\bf 57.63} & {\bf 56.58} &       & {\bf 91.73} & {\bf 91.36} & 90.45 &        \\ \hline
      VLAD+FV             & 85.23 & 84.54 & 83.52 &       & 58.13 & 57.60 & 55.64 &       & 90.91 & {\bf 91.36} & {\bf 90.82} &        \\ \hline
 \end{tabular}
\end{table*}

\emph{D3 is better than VLAD in almost all cases.} In the 12 comparisons between D3 and VLAD, D3 wins in 11 cases. D3 often has a margin even if it use half of number of dimensions of VLAD (\emph{e.g.}, D3 $K=128$ vs. VLAD $K=256$). It shows that the D3 representation is effective in capturing useful information for classification.

\emph{D3 bridges the gap between VLAD and FV.} In practice we often see that FV has higher accuracy than VLAD, but also much higher computational costs. D3 has roughly the same speed as VLAD, but its accuracy is close to that of FV. Compared to VLAD whose accuracies are usually 2--3\% lower than FV, D3 has much closer accuracy rates to FV. On average, D3 is 1\% worse than FV. On the Youtube dataset D3 is better than FV (91.55\% vs. 91.00\%). Given the computational benefits of D3, it can act as an attractive alternative for FV.

\emph{The hybrid D3 / FV representation (nearly) consistently outperforms all other methods.} We show that the hybrid methods are the best performers in Table~\ref{tbl:action}. The D3+FV representation is especially effective: it is the winner in 8 out of 9 cases. With $K=128$ in the Youtube set being the only exception, D3+FV consistently beats other methods, including FV and VLAD+FV.

Two points are worth pointing out. First, the success of D3+FV shows that the information encoded in D3 and FV, although both used first and second order statistics of the two distributions, are complementary to each other. The hybrid of these two outperforms both D3 and FV. Since the running time of D3+FV is only roughly half of that of FV, D3+FV is attractive in both speed and accuracy. Second, VLAD+FV is obviously inferior to D3+FV. Its accuracy is very similar to that of FV, but lower than D3+FV in most cases.

\subsection{Image recognition results} \label{sec:exp_image_deep}

Now we test how D3 (and the comparison methods) work with instance vectors that are extracted by state-of-the-art deep learning methods. To extract instance vectors, we use the DSP (deep spatial pyramid) method~\cite{anonymous}, which spatially integrates deep fully convolutional networks. A set of instance vectors are efficiently extracted, each of which corresponds to a spatial region (\emph{i.e.}, receptive field) in the original image. The CNN model we use is imagenet-vgg-verydeep-16 in~\cite{vi:Simonyan2015} till the last convolutional layer, and the input image is resized such that its shortest edge is no smaller than 314 pixels, and its longest edge is no larger than $1120$ pixels. Six spatial regions are used, corresponding to the level 1 and 0 regions in~\cite{me:Wu2011}. \cite{anonymous} finds that FV or VLAD usually achieves optimal performance with very small $K$ sizes in DSP. Hence, we test $K \in \{4,8\}$.

The following image datasets are used.
\squishlist
 \item Scene 15~\cite{pl:Lazebnik2006}. It contains 15 categories of scene images. We use 100 training images per category, the rest are for testing.
 \item MIT indoor 67~\cite{pl:Quattoni2009}. It has 15620 images in 67 indoor scene types. We use the train/test split provided in~\cite{pl:Quattoni2009}.
 \item Caltech 101~\cite{vi:FeiFei2004}. It consists of 9K images in 101 object categories plus a background category. We train on 30 and test on 50 images per category.
 \item Caltech 256~\cite{vi:Caltech256}. It is a superset of Caltech 101, with 31K images, and 256 object plus 1 background categories. We train on 60 images per category, the rest for testing.
 \item SUN 397~\cite{vi:Xiao2010}. It is a large scale scene recognition dataset, with 397 categories and at least 100 images per category. We use the first 3 train/test splits of~\cite{vi:Xiao2010}.
\squishend

Except for the indoor and SUN datasets, we run 3 random train/test splits in each dataset. Average accuracy rates on these datasets are reported in Table~\ref{tbl:deep_image}. As shown by the standard deviation numbers in Table~\ref{tbl:deep_image}, the deep learning instance vectors are stable and the standard deviations are small in most cases. Thus, we tested with 3 random train/test splits instead of more (\emph{e.g.}, 5 or 10).

\begin{table*}[t]
 \caption{Image recognition accuracy (percent) comparisons. The definition of $K$ is the same as that used in Table~\ref{tbl:action}. The best results are shown in bold face. Standard deviations are also showed after the $\pm$ sign.} \label{tbl:deep_image}
 \centering
 \small
 \begin{tabular}{| l | *{5}{l@{\;\;}l|}  }
     \hline
                & \multicolumn{2}{c|}{Scene 15} & \multicolumn{2}{c|}{MIT indoor 67} & \multicolumn{2}{c|}{Caltech 101} & \multicolumn{2}{c|}{Caltech 256} & \multicolumn{2}{c|}{SUN 397} \\
                &$K=4$&\multicolumn{1}{@{}l|}{$K=8$}&$K=4$&\multicolumn{1}{@{}l|}{$K=8$}&$K=4$&\multicolumn{1}{@{}l|}{$K=8$}&$K=4$&\multicolumn{1}{@{}l|}{$K=8$}&$K=4$&\multicolumn{1}{@{}l|}{$K=8$} \\ \hline 
  D3            & 92.34{\tiny$\pm$0.23} & 92.10{\tiny$\pm$0.65} & 77.31 & 77.76  &  93.60{\tiny$\pm$0.17} & 93.80{\tiny$\pm$0.58} & 83.15{\tiny$\pm$0.15} & 82.92{\tiny$\pm$0.09} & 59.93{\tiny$\pm$0.24} & 60.22{\tiny$\pm$0.07} \\ \hline
  VLAD          & 92.58{\tiny$\pm$0.60} & 92.61{\tiny$\pm$0.42} & \textbf{77.61} & \textbf{78.13}  &  94.20{\tiny$\pm$0.39} & 94.11{\tiny$\pm$0.57} & 84.01{\tiny$\pm$0.02} & 84.00{\tiny$\pm$0.10} & 60.61{\tiny$\pm$0.25} & 61.22{\tiny$\pm$0.33} \\ \hline
  FV            & 91.96{\tiny$\pm$0.40} & 91.53{\tiny$\pm$0.56} & 75.97 & 75.82  &  94.32{\tiny$\pm$0.51} & 94.10{\tiny$\pm$0.33} & 83.75{\tiny$\pm$0.16} & 83.40{\tiny$\pm$0.13} & 58.40{\tiny$\pm$0.12} & 57.97{\tiny$\pm$0.28} \\ \hline \hline
  D3+FV         & \textbf{92.83}{\tiny$\pm$0.55} & 92.82{\tiny$\pm$0.31} & 77.09 & 77.99  &  \textbf{94.72}{\tiny$\pm$0.51} & \textbf{94.51}{\tiny$\pm$0.44} & \textbf{84.77}{\tiny$\pm$0.12} & \textbf{84.62}{\tiny$\pm$0.15} & \textbf{61.48}{\tiny$\pm$0.22} & 61.38{\tiny$\pm$0.52} \\ \hline 
  VLAD+FV       & 92.82{\tiny$\pm$0.52} & 92.76{\tiny$\pm$0.56} & 77.54 & 78.06  &  94.71{\tiny$\pm$0.41} & 94.45{\tiny$\pm$0.51} & 84.18{\tiny$\pm$0.51} & 84.61{\tiny$\pm$0.16} & 61.32{\tiny$\pm$0.26} & \textbf{61.83}{\tiny$\pm$0.27}  \\ \hline
  D3+VLAD       & 92.82{\tiny$\pm$0.30} & \textbf{92.92}{\tiny$\pm$0.19} & 77.01 & 77.91  &  94.59{\tiny$\pm$0.54} & 94.45{\tiny$\pm$0.41} & 84.09{\tiny$\pm$0.25} & 84.31{\tiny$\pm$0.14} & 60.38{\tiny$\pm$0.30} & 61.48{\tiny$\pm$0.32}  \\ \hline \hline
                & \multicolumn{2}{c|}{91.59{\footnotesize$\pm$0.48}~\cite{vi:Zhou2014}} & \multicolumn{2}{c|}{77.56~\cite{vi:Gong2014}} & \multicolumn{2}{c|}{93.42{\footnotesize$\pm$0.50}~\cite{vi:He2014}} & \multicolumn{2}{c|}{77.61{\footnotesize$\pm$0.12}~\cite{vi:Chatfield2014}} & \multicolumn{2}{c|}{53.86{\footnotesize$\pm$0.21}~\cite{vi:Zhou2014}} \\ \hline
 \end{tabular}
\end{table*}

D3 and D3+FV have shown excellent results when combining with instance vectors extracted by deep nets. We have the following key observations from Table~\ref{tbl:deep_image}, which mostly coincides well with the observations concerning action recognition in Table~\ref{tbl:action}. The last row in Table~\ref{tbl:deep_image} shows the current state-of-the-art recognition accuracy in the literature, which are achieved by various systems that depend on deep learning using the same evaluation protocol.

\emph{D3 is slightly better than FV.} D3 is better than FV in 3 datasets (Scene 15, indoor 67 and SUN 397), but worse than FV in the two Caltech datasets. It is worth noting that D3's accuracy is higher than that of FV by a larger margin in indoor 67 (1--2\%) and SUN 397 (1.5--2.2\%), while FV is only higher than D3 by 0.3--0.7\% in the Caltech 101 and 256 datasets. Another important observation is that the win/loss are consistent among the train/test splits. In other words, if D3 wins (loses) in one dataset, it wins (loses) consistently in all three splits.\footnote{Detailed per-split accuracy numbers are omitted.} Thus, the CNN instance vectors lead to stable comparison results, and we believe 3 train/test splits are enough to compare these algorithms.

\emph{VLAD is better than both D3 and FV, but D3 bridges the gap between VLAD and FV.} Although FV usually outperforms VLAD in image classification and retrieval using dense SIFT features and in the action recognition results of Table~\ref{tbl:action}, a reversed trend is shown in Table~\ref{tbl:deep_image} using CNN instance vectors. VLAD is almost consistently better than FV, up to 3.2\% higher in the SUN 397 dataset. The accuracy of D3, however, is much closer to that of VLAD than FV's accuracy. D3 is usually 0.3\%--0.6\% lower than VLAD, with only two cases up to 1.1\% ($K=8$ in Caltech 256 and SUN 397).

\emph{The hybrid methods are all effective, and D3+FV is the overall winning method.} The second part of Table~\ref{tbl:deep_image} presents results of hybrid methods. Beyond D3+FV and VLAD+FV, we also add the results of D3+VLAD, because VLAD is the winner in the first part of Table~\ref{tbl:deep_image}. Excluding the MIT indoor 67 dataset, obviously all hybrid methods have higher accuracy rates than every individual method. Among the hybrid methods, D3+FV is the overall winner again. It has the highest accuracy in 6 cases, while VLAD+FV and D3+VLAD has only one each. When comparing D3+FV with D3, FV or VLAD in detail, this hybrid method has higher accuracy than any single method in all train/test splits in all 36 comparisons (4 datasets excluding the indoor 67 dataset $\times$ 3 individual representations $\times$ 3 train/test splits). The MIT indoor 67 dataset is a special case, where VLAD is better than all other methods. We are not yet clear what characteristic of this dataset makes it particularly suitable for VLAD.

The fact that D3 is in general inferior to VLAD in this setup also indicates that CNN instance vectors have different characteristics than the dense SIFT vectors (cf. Sec.~\ref{sec:exp_property}), for which VLAD is inferior to D3. 

This might be caused by the fact that D3 and VLAD used very small $K$ values ($K=4$ or 8) with CNN instance vectors, compared to $K \ge 64$ in Sec.~\ref{sec:exp_property}. Hence, both methods have much fewer number of dimensions now, and a few VLAD dimensions with highest discriminative powers may lead to better performance than D3. We will leave a careful, more detailed analysis of this observation  to future work.

\emph{Significantly higher accuracy than state-of-the-art, especially in those difficult datasets.} DSP~\cite{anonymous} (with D3 or other individual representation methods) is a strong baseline, which already outperforms previous state-of-the-art in the literature (shown in the last row of Table~\ref{tbl:deep_image}). The hybrid method D3+FV leads to even better performance, \emph{e.g.}, its accuracy is 7.2\% higher than~\cite{vi:Chatfield2014} for the Caltech 256 dataset,\footnote{\cite{vi:Simonyan2015} reported an average recall rate of 86.2\% for Caltech 256. DSP's average recall is 89.12\% and D3+FV is 90.25\% ($K=4$).} and 7.6\% higher than the place deep model of~\cite{vi:Zhou2014} for SUN 397.

\subsection{Discussions} \label{sec:exp_discussions}

Overall, the proposed D3 representation method has the following properties:
\squishlist
 \item \textbf{D3 is discriminative, efficient, and \emph{stable}.} D3 is not the individual representation method that leads to the highest accuracy. FV is the best in our action recognition experiments with ITF instance vectors, while VLAD is the best in our image categorization experiments using CNN features. It is, however, \emph{the most stable one}. It is only slightly worse than FV in action recognition and slightly worse than VLAD in image categorization. Although VLAD is outperformed by FV by a large margin in action recognition (Table~\ref{tbl:action}) and vice versa for image categorization (Table~\ref{tbl:deep_image}), D3 has stably achieved high accuracy rates. D3 is also as efficient as VLAD, and is much faster than the FV method;
 \item \textbf{D3+FV is the overall winning method.} Using the same number of dimensions for all individual and hybrid methods, D3+VLAD has shown the best performance, which indicates that the information encoded by D3 and FV form a synergy. Since the FV part of D3 only uses half the number of Gaussian components than that in individual FV, D3+FV is still more efficient than FV alone.
\squishend
In short, D3 and D3+FV are effective and efficient in encoding entities that are represented as sets of instance vectors.

\section{Conclusions and Future Work} \label{sec:conclusions}

We proposed the Discriminative Distribution Distance (D3) method to encode an entity (which comprises of a set of instance vectors) into a vector representation. Unlike existing methods such as FV, VLAD or Super Vectors that are designed from a generative perspective, D3 is based on discriminative ideas. We proposed to use directional distances to measure how two distributions (sets of vectors) are different with each other, and proposed to use the robust MPM classifier to robustly estimate this distance.

These discriminative design choices lead to excellent classification accuracy of the proposed D3 representation, which are verified by extensive experiments on action and image categorization datasets. D3 is also efficient, and the hybrid D3+FV representation has achieved the best results among compared individual and hybrid methods.

In the same spirit as D3, we plan to combine D3 and FV in a principled way, which will add discriminative perspectives to FV and will further reduce the computational cost of the hybrid representation using D3+FV. We will further study how the benefits of VLAD can be utilized (\emph{e.g.}, when CNN instance vectors are used). 

{\small
\bibliographystyle{ieee}
\bibliography{abbr_s,BibAll}
}

\end{document}